\documentclass[journal,transmag]{IEEEtran}
\usepackage{amsmath,amssymb,amsfonts, amstext}
\usepackage[numbers]{natbib} % for citeauthor
\usepackage{longtable,booktabs}
\usepackage{array}
\usepackage{algorithm}
\usepackage{algpseudocode}
\usepackage{graphicx}
\usepackage{url}
\graphicspath{
  {figs/}
}
\usepackage{lscape} % for the big problem instances table
\usepackage{float}
\usepackage[longtable]{multirow}
\usepackage{listings}
\usepackage{tikz}

\newcommand\mdoubleplus{\mathbin{+\mkern-10mu+}}

\newsavebox\CBox

\newcolumntype{R}{>{$}c<{$}}

\newcolumntype{?}[1]{!{\vrule width #1}}

\begin{document}
\title{Prediction Intervals and Confidence Regions for Symbolic Regression Models based on Likelihood Profiles}


\author{\IEEEauthorblockN{F. O. de Franca\IEEEauthorrefmark{1} and
  G. Kronberger\IEEEauthorrefmark{2}}
\IEEEauthorblockA{\IEEEauthorrefmark{1}Center for Mathematics, Computation and Cognition (CMCC), Heuristics, Analysis and Learning Laboratory (HAL), \\Federal University of ABC, Santo Andre, Brazil (e-mail: folivetti@ufabc.edu.br)}
\IEEEauthorblockA{\IEEEauthorrefmark{2}Josef Ressel Center for Symbolic Regression, \\ Heuristic and Evolutionary Algorithms Laboratory (HEAL),\\ University of Applied Sciences Upper Austria, \\Softwarepark 11, 4232 Hagenberg, Austria}
\thanks{Manuscript received XXXX; XXXX. 
Corresponding author: F. O. de Franca (email: folivetti@ufabc.edu.br}}

\markboth{IEEE TRANSACTIONS ON EVOLUTIONARY COMPUTATION}%
 {de France \MakeLowercase{\textit{et al.}}: Prediction Intervals for Symbolic Regression Models}



\maketitle

% REQUIREMENTS:
% * abstract of 100-200 words --> we now have 135 words
% * no abbreviations
\begin{abstract}
  Symbolic regression is a nonlinear regression method which is commonly performed by an evolutionary computation method such as genetic programming. Quantification of uncertainty of regression models is important for the interpretation of models and for decision making. The linear approximation and so-called likelihood profiles are well-known possibilities for the calculation of confidence and prediction intervals for nonlinear regression models.
  These simple and effective techniques have been completely ignored so far in the genetic programming literature.
  In this work we describe the calculation of likelihood profiles in details and also provide some illustrative examples with models created with three different symbolic regression algorithms on two different datasets. The examples highlight the importance of the likelihood profiles to understand the limitations of symbolic regression models and to help the user taking an informed post-prediction decision.
\end{abstract}


% \fi
\IEEEpeerreviewmaketitle

\section{Introduction}
\label{sec:introduction}

% the problem of randomness in models and introducing intervals
\IEEEPARstart{W}{henever} we build a regression model from data, we are subject to different sources of uncertainties such as measurement noise, choice of explanatory variables, or sampling bias.
% In such situations, it is impossible to have an exact estimate of the predictions made by the model and for the values of the adjustable numerical parameters.
The inferences made by regression models are subject to these uncertainties, having a quantification of such uncertainties can help with a decision process. Imagine a regression model that makes inference about the precipitation at a certain region for a given period of time. Knowing how much it will rain can help warning  people living areas endangered by landslides. Let us say that the decision threshold to move people to shelters is a value of $100$mm of rain or more. If the model predicts $90$mm of rain, the decision would be to keep everyone at their homes. But this number accounts only for the expected values ignoring all the uncertainties involved. By taking the uncertainties into account, we could have reported the same expected value of $90$mm but with a confidence interval $[75, 105]$, for example. With this information we can make a more informed decision so that, if we decide not to move them out of their homes, we can at least keep personnel on alert. 

The same can be said about the parameter estimates; knowing the involved uncertainties can help to interpret the model. Besides inference, another important task of regression analysis is association. In this case, we are interested in understanding the association of a given variable with the output of the system. Suppose we create a linear regression model that investigates the association of education level and height with salary range. If the associated parameters are $1.3, 0.1$ for the respective variables, we can say that increasing education level by $1$ will increase the salary range by $1.3$ units, while increasing your height by one unit, increases the salary range by $0.1$ units. If we accept these values as it is, we would argue that both variables have a positive effect on salary. But, let us say the confidence interval for these coefficients are $[1.0, 1.6]$ and $[-0.1, 0.2]$. In this case, we can be confident that the education level does have a positive effect on salary, but the effect of height may have been simply a random artifact or sample bias.

As can be seen from these examples, having a confidence interval for the predictions and the numerical coefficients can help the practitioner to understand their models and make better decisions. The calculation of confidence intervals is well-established for linear regression with Gaussian likelihood and readily available in most software packages. For nonlinear models, given we know the regression function and its partial derivatives, we can also calculate a linear approximation of the confidence interval~\cite{batesnonlinear1988}. This is frequently called the ``delta method''. Bates and Watts described an even better approach which use the profile $t$ function~\cite{batesnonlinear1988}, also called likelihood profile, to calculate more accurate confidence intervals, with the cost of having to apply a nonlinear solver multiple times. While not as widespread as in linear regression toolboxes, there are implementations of both the linear approximation and likelihood profiles for generalized linear or nonlinear regression models\footnote{The R nls() function is an important example.}.

% move symbolic regression to the beginning of this section because we can early on drop the term "evolutionary algorithm"?
% symbolic reg. and the lack of post-analysis tools
Symbolic Regression (SR) is a well-established regression task~\cite{la2021contemporary}, where the goal is to find the function form and numerical parameters that best fit a certain dataset. It is most commonly solved using genetic programming (GP) ~\cite{koza:book} which is an evolutionary algorithm. Unlike many opaque nonlinear models (e.g., gradient tree boosting, neural networks), a symbolic model can bring a balance between accuracy and interpretability~\cite{Kotanchek:2007:GPTP,bomarito2022automated,Hu:2022:GPTP}. Most modern SR implementations allow for a user tunable control of the desired complexity of the generated model. Also, having a symbolic model, means that we can study the association or effect of the variable to the target using partial effects~\cite{SRPEGecco, SRPEGEMP}. Most importantly SR models are nonlinear regression models and therefore we may use the linear approximation or likelihood profiles to quantify uncertainty of parameter estimates and predictions. Despite this obvious link, likelihood profiles have not been used for SR models and there is no SR tool available today that can provide likelihood profiles and confidence intervals for the predictions and parameters.

% Relevance of this paper.
This paper contains two important contributions. First, we give clear pseudo-code for the calculation of likelihood profiles, approximate contours for pairwise parameter confidence regions, and the calculation of prediction intervals for nonlinear models. Second, we demonstrate how the algorithms can be used in particular for expressions produced by SR tools. The algorithms are implemented in an open-source software module written in Python that can be used for uncertainty quantification of SR models as long as they are compatible with \emph{sympy} Python library.

The concept of likelihood profiles for nonlinear regression is well-known in the statistics community but the algorithms published in \cite{batesnonlinear1988} have some small mistakes which we fixed in this paper. Additionally, the algorithm for profile-based prediction intervals is not described at all in \cite{batesnonlinear1988}. In GP and SR literature uncertainty quantification based on likelihood profiles is completely ignored because the community is seemingly unaware of this technique stemming from the statistics literature. As a consequence SR implementations usually do not provide confidence or prediction intervals at all or use various heuristics such as lower-upper bound estimation (LUBE) ~\cite{Rebuli2022predictioninterval}. Another reason is that the calculation of likelihood profiles is only possible after  symbolic manipulation of SR expressions as we demonstrate in this paper. This requires extra work and can be difficult without the help of computer algebra libraries such as sympy~\cite{sympy}. The Python module published together with this paper solves this issue.

% to calculate the Jacobian w.r.t. the parameters, to rewrite the symbolic expression to calculate the prediction intervals, and to simplify some common constructs to avoid multicolinearity among the coefficients, which can lead to a large error for the coefficients estimation. While the first requirement can be straightforward to implement, and can be already automated with the use of libraries such as \emph{sympy}~\cite{} or automatic differentiation libraries, the other two require some extra work.

The paper is organized as follows: we first discuss related work in Sec.~\ref{sec:related-work} and Sec.~\ref{sec:ci} details the calculation of CI in linear regression, the linear approximation for nonlinear models and the use of the likelihood profile for nonlinear intervals. In Sec.~\ref{sec:srprofile} we provide the requirements and algorithmic details of the CI calculation for symbolic regression models.
Following up, in Sec.~\ref{sec:demo} we demonstrate the use of the algorithms. Finally, in Sec.~\ref{sec:conclusion} we discuss limitations of the approach and some additional insights as well as future steps.

\subsection{Related Work}
\label{sec:related-work}
% Any prediction model is more useful when the model also provides information on the certainty of predictions. Additionally, in interpretable models the certainty of parameter estimates is valueable. Correspondingly,
Several techniques have been described for calculating prediction or confidence intervals for SR models.

GP using intervals instead of crisp values has been used together with interval arithmetic to produce prediction intervals in~\cite{sanchez:2000:TEC}, but the results were not compared to the delta method or profile-based intervals. Later, the conformal prediction technique~\cite{conformalprediction2014} developed in the machine learning domain was used for GP instead of interval arithmetic in~\cite{Thuong:2017:GECCO}.

Ensembles of GP models are used for calculating prediction uncertainty in~\cite{Parasuraman:2008:WRR}. This work is notable because it not only accounts for the uncertainty of parameters but also for the uncertainty of model structures identified by GP. This second type of uncertainty is ignored by the traditional methods which focus only on the uncertainty of parameter estimates. A similar approach based on Bayesian model averaging is described in~\cite{Agapitos:2014:CEC}. Such ensembling methods can be improved even more by using the methods described in this paper to incorporate the prediction uncertainty of the individual models in the ensemble. Such an approach has been described recently in~\cite{Werner2022uncertainty}. Here the local uncertainty of individual models and global uncertainty from different structures produced by GP is combined using a Bayesian approach. For the calculation of local uncertainty a Laplace approximation is used around maximum a-posteriori parameter values which is conceptually related to the delta method for likelihood-based models. Global uncertainty is determined through Bayesian model averaging.

The lower-upper-bound-estimation (LUBE) method developed originally for neural networks was used together with GP for the particular application of wind speed forecasting~\cite{Li2021}. LUBE is another heuristic approach developed in the area of machine learning to approximate prediction intervals. It was compared to the delta method and Bayesian model selection for neural networks in~\cite{Khosravi2011} and was second only to Bayesian model selection and produced better results than the delta method. LUBE-GP is proposed in~\cite{Rebuli2022predictioninterval} where a multi-objective GP approach is used to minimize prediction interval width and maximize probability coverage. This work is another example that completely ignores the delta method and likelihood profiles.

Another Bayesian approach for uncertainty quantification for SR and GP is described in~\cite{bomarito2022automated}, where the main focus is on using the uncertainty for Bayesian model selection. The Bayesian approach has the advantage that the full posterior is taken into account including especially  multiple locally optimal parameterizations. This is in contrast to the methods proposed in this work which only consider the local neighbourhood around the estimated parameters. A drawback of Bayesian approaches is the requirement to define a model prior and the computational effort required for sampling from the posterior.

% XXX There is also a recent paper by Trujillo (so far unpublished but we should check arxiv.org before we submit.

% paper objective and organization
% F: remove this paragraph?
% G: yes
% As such, this paper has the objective of filling this gap and to propose a generic algorithm to calculate the CI of predictions and numerical coefficients for symbolic models using the linear approximation and the profile $t$ function. As a byproduct of this proposal, we have made available a Python module that, given a symbolic model, automates all the process and provides a standard textual report and plots of interest for regression analysis. We also provide some example usages integrating this script with HeuristicLab, TIR~\cite{tir}, and deterministic symbolic regression and show that it is straightforward to adapt to other SR toolboxes as well.

\section{Confidence Intervals for Nonlinear Regression Models}
\label{sec:ci}
In the following we summarize the basics of calculating confidence intervals for parameters and prediction intervals for nonlinear regression models using the linear approximation as well as the likelihood profile method as described in \cite{batesnonlinear1988}.

% details of linear CI and algorithm for calculation
Given a dataset formed by $n$ samples of $p$ independent variables $\mathbf{X} \in \mathbb{R}^{n \times p}$ and corresponding dependent (also called target) measurements $\mathbf{y} \in \mathbb{R}^n$, a linear regression model parameterized by numerical coefficients $\beta \in \mathbb{R}^p$ is given by:

\begin{equation}
    \mathbf{y} = \mathbf{X} \beta + \mathbf{\epsilon},
\end{equation}
where $\mathbf{\epsilon}$ is called residual and corresponds to the error term associated with the uncertainty of the collected data. The linear model assumes that the residuals are i.i.d. normally distributed with zero mean and that the relationship between the independent and dependent variables is linear.

The $X \beta$ part of the model is called the expectation function and it is related to the inference of the expected value of the target variable ($\mathbf{E}[\mathbf{y}] = \mathbf{X} \beta$). Specifically for the linear model, $\mathbf{X}$ is also called the derivative matrix, as the partial derivative of $\mathbf{y}$ w.r.t. $\beta$ is $\mathbf{X}$.

To determine the values of $\beta$ we can solve the least square problem by minimizing the residual sum of the squares ($\mathit{SSR}$):

\begin{equation}
  SSR(\beta) = \| \mathbf{y} - \mathbf{X} \beta \|_2^2.
\end{equation}
The solution to this optimization problem is simply:

\begin{equation}
   \hat{\beta} = (\mathbf{X}^T \mathbf{X})^{-1}\mathbf{X}^T \mathbf{y}.
\end{equation}

The residual mean square ($s^2$) or variance estimate is calculated based on $n - p$ degrees of freedom:

\begin{equation}
    s^2 = \frac{\mathit{SSR}(\mathbf{\hat{\beta}})}{n-p}.
\end{equation}

The $1 - \alpha$ confidence interval of a coefficient $\beta_i$ can be calculated using the upper $\alpha/2$ percentile for a $t$-distribution with $n-p$ degrees of freedom:

\begin{equation}
    \hat{\beta_i} \pm \mathit{se}(\hat{\beta_i})\, t(n-p, \alpha/2),
   \label{eq:beta_ci}
\end{equation}
where

\begin{equation}
    \mathit{se}_i = s\sqrt{((\mathbf{X}^T\mathbf{X})^{-1})_{ii}},
\end{equation}
where $A_{ii}$ represents the $i$-th diagonal element of a matrix $A$. Similarly we can calculate the $1 - \alpha$ confidence interval of the prediction at a new point $\mathbf{x}$ as:

\begin{equation}
   \mathbf{x}^T\mathbf{\hat{\beta}} \pm s \sqrt{\mathbf{x}^T(\mathbf{X}^T\mathbf{X})^{-1}\mathbf{x}}\, t(n-p, \alpha/2).
   \label{eq:y_ci}
\end{equation}

% problems for nonlinear models
When working with nonlinear models, we are interested in models:

\begin{equation}
    \mathbf{y} = f(\mathbf{X}, \mathbf{\theta}) +\mathbf{\epsilon}, 
\end{equation}
where $\mathbf{\theta}$ corresponds to the numerical coefficients of the model and we assume a probability distribution for $\mathbf{\epsilon}$ (often a Gaussian distribution). This allows us to estimate $\mathbf{\theta}$ using the maximum likelihood method which in this case requires iterative nonlinear optimization algorithms. These algorithms update the coefficients in the opposite direction of the gradient. One example is the Gauss-Newton method that updates an initial $\mathbf{\theta}_0$ with the linearization:

\begin{equation}
  \mathbf{\theta}_{t+1} = \mathbf{\theta}_t - (J^T J)^{-1} J^T \epsilon_t,
\end{equation}
where $J$ is the Jacobian matrix for $n$ observations evaluated for the current point $\mathbf{\theta}_t $ such that

\begin{equation}
  J_{kj} = \frac{\partial \epsilon_k}{\partial \theta_j},\, k=1\ldots n.
\end{equation}

The iteration is resumed until convergence to a local optimum which can be measured by the size of the parameter increment. The local optimum provides the maximum likelihood estimate for the parameters $\hat{\mathbf{\theta}}$. 

\subsection{Linear approximation}
Now, given the $\mathit{QR}$ decomposition of $J$ evaluated for $\hat{\mathbf{\theta}}$, we can calculate the linear approximation for the standard error $\mathit{se}_i$ of the $i$-th parameter as the length of the $i$-th row of $R^{-1}$

\begin{equation}
    \mathit{se}_i = s \sqrt{ \sum_j{(R^{-1})_{ij}^2} }.
\end{equation}

The residual standard error ($\mathit{rse}$) can be calculated as:

\begin{equation}
    \mathit{rse}_k = s \sqrt{ \sum_j{(J R^{-1})_{kj}^2} },
\end{equation}
where $J$ is evaluated for new points and $\hat{\mathbf{\theta}}$ and $R^{-1}$ is for the training data $\mathbf{X}$ and $\hat{\mathbf{\theta}}$. 

The confidence intervals can be calculated the same way as in Eq.~\ref{eq:beta_ci}. The prediction interval for the expectation function for any point $\mathbf{x}$ can be calculated as:
\begin{equation}
  f(\mathbf{x}, \hat{\mathbf{\theta}}) \pm \mathit{rse} \, t(n-p, \alpha/2). 
\end{equation}
The prediction interval for the full model including the noise term is:
\begin{equation}
  f(\mathbf{x}, \hat{\mathbf{\theta}}) \pm (\mathit{rse} + s) \, t(n-p, \alpha/2). 
\end{equation}
% we only show pointwise intervals and not confidence bounds (simultaneous intervals)

Notice, though, that these confidence intervals are a linear approximation of the true intervals and can be ``extremly misleading''~\cite{batesnonlinear1988}.

\subsection{Likelihood profile}

The likelihood profile \cite{batesnonlinear1988} gives more accurate confidence intervals for nonlinear regression models  -- though without the guarantee of being exact, as pointed out in~\cite{quinn2000notes}. This technique uses the likelihood profile  $\tau(\theta_i)$ for the $i$-th coefficient defined as:

\begin{equation}
    \tau(\theta_i) = \frac{\mathit{sign}(\theta_i - \hat{\theta}_i)}{s} \sqrt{ \overline{\mathit{SSR}}(\theta_i) - \mathit{SSR}(\mathbf{\hat{\theta}}) },
\label{eq:tau}
\end{equation}
where $\overline{SSR}(\theta_i)$ is the optimized sum of squared residuals obtained where the value of $\theta_i$ is held fixed and all other parameters re-optimized starting from $\mathbf{\hat{\theta}}$.
In contrast to Eq.~\ref{eq:beta_ci}, the CI of $\theta_i$ is the set of all values such that:

\begin{equation}
  -t(n - p, \alpha/2) \leq \tau(\theta_i) \leq t(n - p, \alpha/2)
\end{equation}

In the following we summarize the algorithms described in~\cite{batesnonlinear1988}.
To create the function $\tau(\theta_i)$ we sample a set of points of $\tau(\hat{\theta}_i + \delta)$ for different values of $\delta$. This is done by setting $\hat{\theta}_i = \hat{\theta}_i + \delta$ and optimizing the least squares problem while keeping the $i$-th coefficient fixed. After that, we calculate the corresponding value of $\tau$ following Eq.~\ref{eq:tau}. To determine the points to be sampled, we start with a value of $\delta = \mathit{se}_i/ \mathit{step}$ where $\mathit{step} = 8$ the initial step size. After calculating a value of $\tau$ we determine the next step based on the slope of $\tau$ and step $t$ limiting to a maximum of $k_{max} = 30$ points or reaching a maximum absolute value of $\sqrt{f(1 - 0.01, p, n - p)}$ (i.e., the limit of $\tau$ for a $99\%$ confidence interval). We do the same for decreasing $\delta$ values starting from $\delta = -\mathit{se}_i / \mathit{step}$ and finally add the trivial point $\tau(\hat{\theta}_i) = 0$. In Alg.~\ref{alg:profilet} we describe the whole process in detail. One caveat with this procedure is that if, for any reason, the optimization procedure finds a better optimum, the current $\hat{\mathbf{\theta}}$ should be replaced by the new one and the process must be restarted from the beginning with the new parameters values. This can happen if the optimization method was interrupted before convergence (i.e., reached maximum iteration) or if the disturbance caused by $\delta$ leads to a new basin of attraction with a better optimum.

\begin{algorithm}
\begin{algorithmic}[1]
\Function{Profile}{$i, \hat{\mathbf{\theta}}, f, J$}
  \State $T \gets [ ]$
  \State $\Theta \gets [ ]$
  \State $\tilde{\theta} \gets \theta$
  \ForAll{$\delta \in [\frac{-\mathit{se}_i}{\mathit{step}}, \frac{\mathit{se}_i}{\mathit{step}}]$}
  \State $\mathit{invSlope} \gets 1$
  \State $t \gets 1$
    \For{$k = 1 \ldots k_{max}$}
       \State $\tilde{\mathbf{\theta}}_i \gets \theta_i + \delta t$
       \State $\tilde{\mathbf{\theta}} \gets \mathit{nls}(\tilde{\mathbf{\theta}}, f, \tilde{J})$ %\Comment{$\tilde{J})$ is the Jacobian with the $i$-th column replaced by $0$.}
       \State $\tau \gets \frac{\mathit{sign}(\theta_i - \hat{\theta}_i)}{s} \sqrt{ \overline{\mathit{SSR}}(\theta_i) - \mathit{SSR}(\mathbf{\hat{\theta}}) }$
       \State $T \gets T \mdoubleplus [\tau]$
       \State $\Theta \gets \Theta \mdoubleplus [\tilde{\mathbf{\theta}}]$       
       \State $\mathit{invSlope} \gets \left|\frac{\tau s^2}{\mathit{se}_i \epsilon^T J_{\_i}}\right|$ %\Comment{$\epsilon$ is the residue and $J_{\_i}$ is the $i$-th column of the Jacobian.}
       \State $\mathit{invSlope} \gets \min(4, \max(\mathit{invSlope}, 1/16))$
       \State $t \gets t +\mathit{invSlope}$
       \If{$|\tau| > \tau_{max}$}
            \State break
       \EndIf 
    \EndFor
  \EndFor
  \State $T \gets T \mdoubleplus [0]$
  \State $\Theta \gets \Theta \mdoubleplus [\hat{\mathbf{\theta}}]$
  \State \Return $T, \Theta$
\EndFunction
\end{algorithmic}
\caption{Likelihood profile algorithm. In this algorithm $\tilde{J}$ is the Jacobian with the $i$-th column replaced by $0$, $\epsilon$ is the residue for $\tilde{\mathbf{\theta}}$, $J_{\_i}$ is the $i$-th column of the Jacobian, and $\mdoubleplus$ is a list concatenation operator. }
\label{alg:profilet}
\end{algorithm}

With the sampled points for $\tau(\theta_i)$ we create the cubic splines $\lambda_{\tau_i \rightarrow \theta_i}, \lambda_{\theta_i \rightarrow \tau_i}$ to interpolate $\tau \rightarrow \theta_i$ and $\theta_i \rightarrow \tau$ and use the former to find the corresponding values of $\theta_i$ for the boundaries $[-t(n - p, \alpha/2), t(n - p, \alpha/2)]$. The latter function is used to create the profile plot of $\tau(\theta_i)$ over $\theta_i$. This plot is insightful to verify the nonlinearity of the coefficient.  %One example of such plot is given in Fig.~\ref{fig:theta_tau} where we can see the difference between the linear approximation and the profile t function.

Bates and Watts also sketch an algorithm for the approximation of pairwise parameter confidence regions based on the likelihood profiles \cite{batesnonlinear1988}, described in Alg.~\ref{alg:thetathetaplot}.
The contours produced by this algorithm are good approximations for the true confidence regions when the parameter estimates are well-determined and close to ellipsoid as shown in the example plots below.
% This plot reveals the confidence regions for pairs of coefficients.
% Another interesting plot is the pair-plot of the confidence interval of two coefficients.% We can build an interpolation of this plot using Alg.~\ref{alg:thetathetaplot}. We will illustrate such plots in Section~\ref{sec:demo}.% An example of such plot is given in Fig.~\ref{fig:thetatheta}, we can see from this figure how ... % F: will insert figure later

\begin{algorithm}
\begin{algorithmic}[1]
\Function{PrepareSplines}{$i, j, \Theta, T, \tau_{scale}$}
  \State $g_{ij} \gets arccos(\lambda_{\theta_i \rightarrow \tau_i}(\Theta_j) / \tau_{scale})$
  \State $\lambda_{\tau_j \rightarrow g_{ij}} \gets $ \Call{CubicSpline}{$\tau_j, g_{ij}$}
  \State \Return $\lambda_{\tau_j \rightarrow g_{ij}}$
\EndFunction
\Statex
\Function{ProfileContour}{$i, j, \Theta, T, \alpha, steps$}
  \State $\tau_{scale} \gets \sqrt{p\, F(1 - \alpha, p, n-p) }$
  \State $\lambda_{\tau_j \rightarrow g_{ij}} \gets $ \Call{PrepareSplines}{$i, j, \Theta, T, \tau_{scale}$}
  \State $\lambda_{\tau_i \rightarrow g_{ji}} \gets $ \Call{PrepareSplines}{$j, i, \Theta, T, \tau_{scale}$}
  \State $\mathit{angle}_0 \gets (0, \lambda_{\tau_i \rightarrow g_{ij}}(1))$
  \State $\mathit{angle}_1 \gets (\lambda_{\tau_i \rightarrow g_{ji}}(1), 0)$
  \State $\mathit{angle}_2 \gets (\pi, \lambda_{\tau_i \rightarrow g_{ij}}(-1))$
  \State $\mathit{angle}_3 \gets (\lambda_{\tau_i \rightarrow g_{ji}}(-1), \pi)$
   \For{$k = 0 \ldots 3$}
   \State $a_k \gets (\mathit{angle}_{k0} + \mathit{angle}_{k1})/2$
   \State $d_k \gets \mathit{angle}_{k0} - \mathit{angle}_{k1}$
   \State $a_k \gets \mathit{sign}(d_k)a_k$
   \State $d_k \gets \mathit{sign}(d_k)d_k$
   \EndFor
   \State $a_4 \gets a_0 + 2\pi$
   \State $d_4 = d_0$
\State $\lambda_{a \rightarrow d} \gets $ \Call{PeriodicCubicSpline}{$a, d$}
\For{$k = 1 \ldots \mathit{steps}$}
\State $x \gets 2k \pi / (\mathit{steps }- 1) - \pi$
      \State $y \gets \lambda_{a \rightarrow d}(x)$
      \State $\tau_{ik} \gets \cos(x + y/2) \tau_{scale}$
      \State $\tau_{jk} \gets \cos(x - y/2) \tau_{scale}$
      \State $\theta_{ik} \gets \lambda_{\tau_i \rightarrow \theta_i}(\tau_{ik})$
      \State $\theta_{jk} \gets \lambda_{\tau_j \rightarrow \theta_j}(\tau_{jk})$
   \EndFor
   \State \Return $\theta_{i}, \theta_{j}$
   \EndFunction
\end{algorithmic}
\caption{Algorithm for the calculation of approximated contour plots for the pairwise parameter confidence regions. This algorithm assumes the existence of a function called \emph{CubicSpline} that returns a cubic splines function based on the input points. \emph{PeriodicCubicSpline} returns the spline with period $2\pi$. $F(1-\alpha, p, n-p)$ is the critical value for the F-distribution with $p$ and $n-p$ degrees of freedom.}
\label{alg:thetathetaplot}
\end{algorithm}

\subsection{Profile-based prediction intervals}
% Prediction CI - rewriting trick
Likelihood profiles can also be used to calculate nonlinear prediction intervals. For this purpose we apply the same algorithm described in Alg.~\ref{alg:profilet} but we have to re-parameterize the model as described in the following. The idea is to re-parameterize the model so that one of the parameters is the output of the model at the evaluation point $x_0$. Let us suppose the nonlinear model is described as:

% here the matrix notation is not necessary and potentially confusing because we do this for each point.
\begin{equation}
  \hat{y} = f(x) = \hat{\theta}_0 e^{\hat{\theta}_1 x}.
  \label{eqn:model}
\end{equation}

To calculate the prediction interval for $\hat{y}$ at point $x_0$, we first rearrange the equation to extract one of the coefficients. Let us pick the first coefficient, then we will have:

\begin{equation}
  \hat{\theta}_0 = \hat{y} e^{-\hat{\theta}_1 x}.
\end{equation}

We then rename $\hat{y}$ in the above expression as our new parameter $\theta_0'$ which represents the output of the model in point $x_0$ and replace in the expectation function:

\begin{equation}
  f(x)' = \theta_0' e^{-\hat{\theta}_1 x_0} e^{\hat{\theta}_1 x}.
  \label{eq:rewrit} 
\end{equation}

The value for the new parameter $\theta_0'$ is calculated via Eq.~\ref{eqn:model} for the point $x$.

Now, we apply Alg.~\ref{alg:profilet} with the re-parameterized model $f(x)'$ for the new parameter $\theta_0'$ after calculating the standard errors for the re-parameterized model whereby we use the training set $X$ for the optimization.
% Additionally, line $10$ is replaced with $\tilde{\mathbf{\theta}}_0 \gets \hat{y}_i + \delta t$ and in line $18$, $\mathit{se}_i$ is replaced by $\mathit{rse}_i$. Everything else remains the same.

These steps ensure that one of the model parameters equals the prediction $\hat{y}_i$ at the point $x_0$ which can be easily verified in Eq.~\ref{eq:rewrit}. The likelihood profile for this parameter provides the nonlinear prediction interval in this point. The re-parameterization and profile calculation has to be repeated for each point for which we evaluate the prediction interval.

\section{Likelihood Profiles for Symbolic Regression}
\label{sec:srprofile}

The algorithms described in the previous section are applicable to any regression model with parameters fitted using the least squares method. The approach is therefore applicable to models produced by SR when the numerical parameters are optimized after the model is generated. This is done in several state-of-the-art implementations for SR~\cite{Burlacu:2020:GECCOcomp,kommenda2020parameter,tir}. 
%a least squares approach is used. A requirement is that the SR model parameters are optimized using nonlinear least squares. This is done in several state-of-the-art implementations for SR~\cite{Kommenda,Burlacu}. 

Another requirement for the applicability is that the parameters of the model are well-behaved. In particular the models should not contain linearly dependent parameters. Models produced by GP systems often do not fulfill this requirement~\cite{Kronberger2022synasc} and have to be simplified before they can be used for the likelihood profile calculation.

% To calculate likelihood profiles for SR models, we need to be capable of performing algebraic manipulation and symbolic computation into the generated expressions.

We assume that the SR implementation produced a model as an expression which contains the optimized parameters. This model may for instance be given as a Python expression that we first have to parse into a symbolic form. For the symbolic representation we have to create:

% Given data points $X, y$ and a symbolic model, we need to create:
% 
\begin{itemize}
   \item A new symbolic expression replacing all numeric values with parameter variables $\theta$.
   \item An evaluator function that receives $\theta$ and inputs $x$ as argument and returns the predictions for the data points.
   \item An evaluator function that receives $\theta$ and inputs $x$ as argument and returns the associated Jacobian matrix.
   \item A function that rewrites the original expression following Eq.~\ref{eq:rewrit}. This should also provide a function that returns the Jacobian matrix for the re-parameterized model.
 \end{itemize}
The first item is straightforward and the second item follows from the first using a recursive tree evaluation function.

Having an expression tree containing only differentiable functions, it is possible to create the symbolic derivative for each variable, thus creating evaluators to compose the Jacobian matrix.
For the last item we rely on \emph{sympy}. % with a stateful variable storing the current unused parameter index. Whenever the algorithm visits a numerical node, it replaces with a variable node named after the current index.

% this is already 
%One possible problem when evaluating CI for SR models is when two or more numerical parameters have a high correlation leading to a rank deficient Jacobian that induces a large value for the standard errors. This can be common in particular for SR implementations since they often apply linear scaling~\cite{Keijzer} of the model. Consider the following symbolic expression:
As already mentioned, one possible issue is when two or more parameters are linearly dependent, leading to ill-conditioning during the optimization process. Consider the following symbolic expression:

\begin{equation}
  f(x) = 0.32 + \frac{5.14}{3x + 1}.
\end{equation}

Replacing the numerical variables to parameters, we have:

\begin{equation}
  f(x) = \theta_0 + \frac{\theta_1}{\theta_2 x + \theta_3}.
\end{equation}

In this particular expression, $\theta_1$ correlates with $\theta_2, \theta_3$, making the problem ill-defined which implies that it is impossible to calculate a reasonable CI for the parameters. Even the linear approximation will have overestimated intervals due to the large standard errors. Unfortunately, GP systems tend to produce such overparameterized expressions.

%We therefore have to remove such redundant parameters before calculating likelihood profiles.
%One way to solve this issue is, during equation rewriting, to detect whenever we see a pattern of product of a number to the sum of symbolic terms (when every term itself has a multiplicative coefficient), we do not replace the outer number
To solve this issue we have to apply rewriting rules to remove linearly dependent parameters. While this can demand advanced symbolic manipulation algorithms, we can apply simple rules for frequent patterns. For instance, whenever there is any part of the expression following the form $\theta_i (\sum_{j,k}{\theta_j x_k})$, we simply fix the value of $\theta_i$ so it is no longer an adjustable parameter. Manual interaction may be required to remove all linear dependencies. This task is easily accomplished by checking the standard errors of parameter estimates and the parameter correlation matrix.

% if the above is clear, we can remove this example.
In our example, we would have the following expression:

\begin{equation}
  f(x) = \theta_0 + \frac{5.14}{\theta_1 x + \theta_2}.
\end{equation}
Alg.~\ref{alg:symmodel} describes this process in details returning the original values of $\theta$ and the symbolic model with all numerical values replaced with parameters variables. % maybe we can drop this algorithm in exchange of space.

\begin{algorithm}
\begin{algorithmic}[1]
%\Function{ExistsNumberAdd}{$\mathit{nodes}$}
%\State $\mathit{hasNumber} \gets \mathit{False}$
%\State $\mathit{hasAdd} \gets \mathit{False}$
%\For{$\mathit{node} \in \mathit{nodes}$}
%  \If{\Call{isNumber}{$node$}}
%    \State $\mathit{hasNumber} \gets \mathit{True}$
%  \EndIf
%  \If{\Call{isAdd}{$node$}}
%    \State $\mathit{hasAdd} \gets \mathit{True}$
%  \EndIf
%\EndFor
%\State \Return $\mathit{hasNumber} \land \mathit{hasAdd}$
%\EndFunction
%
%\Statex 
%
\Function{Rewrite}{$expr, i$, can\_replace}
\If{\Call{isNumber}{$expr$}}
    \If{can\_replace}
    \State \Return $[expr]$, \Call{Symbol}{$\theta_i$}, $i+1$
    \Else
    \State \Return $[], \mathit{expr}, i$
    \EndIf
  \EndIf
  \If{\Call{isSymbol}{$expr$}}
  \State \Return $[], \mathit{expr}, i$
  \EndIf
\State $\mathit{children} \gets $ \Call{childsOf}{$expr$}
\State $\mathit{args} \gets []$
\State $\mathit{values} \gets []$
\For{$\mathit{child} \in \mathit{children}$}
\If{\Call{isNumber}{$\mathit{child}$} $ \land $ \Call{MulAddPat}{$\mathit{expr}$}}
   \State $vals, arg, i \gets $ \Call{Rewrite}{$child, i, False$}
\Else
   \State $vals, arg, i \gets $ \Call{Rewrite}{$child, i, True$}
\EndIf
    \State $values \gets values + vals$
    \State $args \gets args + [arg]$
  \EndFor
   \State $expr \gets $ \Call{replaceChildren}{$expr, args$}
   \State \Return $values, expr, i$
   \EndFunction
\end{algorithmic}
\caption{Algorithm for symbolic rewriting of SR expressions. Numbers are replaced by parameters and linear parameters are removed. The function \emph{MulAddPat} returns true if the current node of the tree represents a multiplication of addition pattern as described on the text.}
\label{alg:symmodel}
\end{algorithm}

\section{ProfileT Python library}

% We must change this once we are 100% of which ones are we using. Also, if using HL we must remove the "as they provide python library"
%As a proof of concept, we have tested our implementation with the expressions produced by two SR frameworks. We have chosen Operon~\cite{Burlacu:2020:GECCOcomp} and TIR~\cite{tir} as they both provide a Python library that returns the symbolic model as a string. The algorithm can be used for SR implementations produce the identified model in the form of a Python expression.
The \emph{ProfileT} Python library provides the \emph{ProfileT} and \emph{SymExpr} classes implementing the likelihood profile calculation for nonlinear regression models described as a \emph{sympy}-compatible \emph{string}.
This library is freely available at \url{https://github.com/folivetti/profile_t} with a full documentation and some usage examples.

\subsection{Demonstration}
\label{sec:demo}

% For the demonstration we use only univariate or bivariate models because it is easy to visualize the prediction intervals for such models. However, the approach is applicable to higher-dimensional datasets.

To demonstrate the calculation of likelihood profiles and prediction intervals, we use three SR implementations: HeuristicLab (HL), Transformation-Interaction-Rational (TIR), and Deterministic Symbolic Regression using Grammar Enumeration (DSR), each of which implements a different algorithm for SR. HeuristicLab uses tree-based GP very similar to Koza-style GP~\cite{Kommenda2012}, TIR uses a restricted model structure and an evolutionary algorithm in combination with ordinary least squares~\cite{tir}, and
DSR uses a tree search algorithm using a formal grammar that restricts model structures~\cite{KammererGPTP}. Coefficients are optimized using the Levenberg-Marquardt algorithm.% FFX (fast function extraction) is a deterministic algorithm which identifies a sparse linear model from a large set of basis functions using LASSO regression for sparsification~\cite{McConaghy:2011:GPTP}.

We use the \emph{PCB} dataset from~\cite{batesnonlinear1988}  which contains measurements of the ``concentration of polychlorinated biphenyl residues in a series of lake trout from Cayuga Lake, NY''. As in~\cite{batesnonlinear1988} we use the logarithm of the concentration as the target variable. We use this dataset because it can be described with a very simple model that is easy to understand. For brevity, we will only show the model generated by HL.
Additionally we used the \emph{Kotanchek} function (Eq.~\ref{eqn:kotanchek}) as in ~\cite{Vladislavleva:2009:TEC,smits:2004:GPTP} for which we will compare models produced by all three SR implementations.
We chose these two data sets since they are both nonlinear and have only one (PCB) or two variables (Kotanchek), making it easier to visualize the prediction intervals. 

% We chose this function because it is frequently used in SR literature, is nonlinear and only has two input variables which %makes it easier to visualize prediction intervals.

% also interesting:
% Lubricant (``kinematic viscosity of a lubricant''), Chlorid (``rate of transport of sulfite ions from blood cells suspended in a salt solution''), and Nitride Utilization (``utilization of nitride in bush beans as a function of light intensity'').
% We use all observations for model training for these two datasets.

% Additionally, we use three benchmark functions from SR literature: \emph{Kotanchek}~\ref{eqn:kotancheck}~\cite{Vladislavleva:2009:TEC,smits:2004:GPTP}, \emph{RatPol2D}~\ref{eqn:ratpol2d}~\cite{Vladislavleva:2009:TEC}, and \emph{Pagie}~\ref{eqn:pagie}~\cite{pagie97evolutionary}. The data for training and test sets are generated as reported in Table~\ref{tab:sampling}. 

%%%%%
% PCB
%%%%%

%% TODO: the text flow needs to be improved.

\subsection{Analysis of SR likelihood profiles for the PCB dataset}

The model for $\log(\mathit{PCB})$ identified by HL is
%EXP(-0.190294172331471 * age) * -3.92922707210363 + 3.12932115590722
\begin{equation}
  -3.93 \exp(-0.19\, \text{age}) + 3.13\,
  \label{eqn:hl-logpcb}
\end{equation}
with $s^2=0.247$ and 3 parameters. This is slightly worse than the handcrafted model in \cite{batesnonlinear1988} which has $s^2=0.246$ with only two parameters.
% The book function also has more plausible extrapolation and a predicted value closer to zero for age=0.
We used SR with shape-constraints as described in \cite{kronberger2022shape} to enforce that the model is smoothly monotonically increasing over age similarly to the reference model. Without these constraints HL produced implausible and overly complex models.
The resulting model cannot be further simplified and has no redundant parameters. 

\begin{figure}
  \centering
  \input{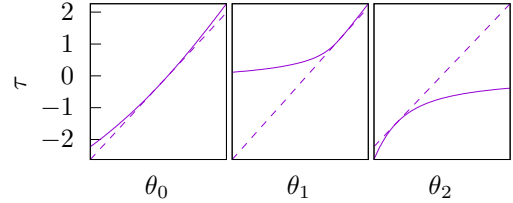}
  \caption{Likelihood profiles for  Eq.~\ref{eqn:hl-logpcb} using linear approximation (dashed line) and the likelihood profile (solid line). We can see from this plot that, for this model, the parameters strongly deviate from linearity.}
  \label{fig:profile-logpcb}
\end{figure}

Fig.~\ref{fig:profile-logpcb} shows the likelihood profiles for each parameter. In this plot the $x$-axis is the parameter value and the $y$-axis is the corresponding value of $\tau$, the value of $\tau$ is related to the degree-of-confidence so, in this particular plot, the values of $\theta$ that corresponds to a $-t(n-p, 0.99/2) < \tau < t(n-p, 0.99/2)$ are the values for a $99\%$ confidence interval. For positive values of $\tau$, if the profile t plot is above the linear approximation or if it is below the linear approximation when $\tau$ is negative, it means it will have a tighter bound than the linear approximation to the right or left. We can see that for $\theta_0$ the left and right bounds are smaller than the linear approximation. In some cases it can also reveal that part of the interval is unbounded, as we can see in the middle and right plots. In these plots the construction of any marginal likelihood extends to $-\infty$ and $\infty$, as there is no corresponding values of $\theta$ for positive and negative counterparts of $\tau$. In some situations the intervals are bound only for a certain likelihood interval.

\begin{figure}
  \centering
  \input{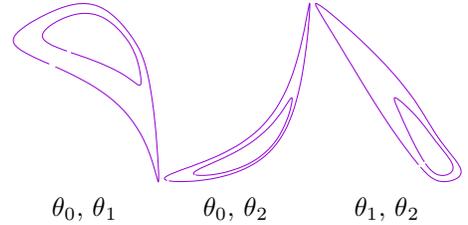}
  \caption{Pairwise approximate parameter confidence regions for  Eq.~\ref{eqn:hl-logpcb}. The outer contour shows the 80\% region, the inner contour the 50\% region.}
  \label{fig:pairwise-logpcb}
\end{figure}

Fig.~\ref{fig:pairwise-logpcb} shows the pairwise likelihood of each pair of parameters. In this plot we can see how the parameters interact.
% For example, given the nonlinearity of the parameters, for a given value of $\theta_0$, the interval of $\theta_1$ is narrower than the ones given by the independent analysis (as in Figure~\ref{fig:profile-logpcb}).
When the parameters are close to linear, the pairwise plots will be ellipsoid, larger correlations leading to elongated ellipses. Strong nonlinearity leads to ``banana-shaped'' contours.

%The likelihood profiles for the parameters shown in Figure~\ref{fig:profile-logpcb} highlight the strong nonlinearity of the parameters around the maximum likelihood estimate. Accordingly, the approximations for the pairwise confidence regions of the parameters shown in Figure~\ref{fig:pairwise-logpcb} are far from ellipsoid.

\begin{figure}
  \centering
  \input{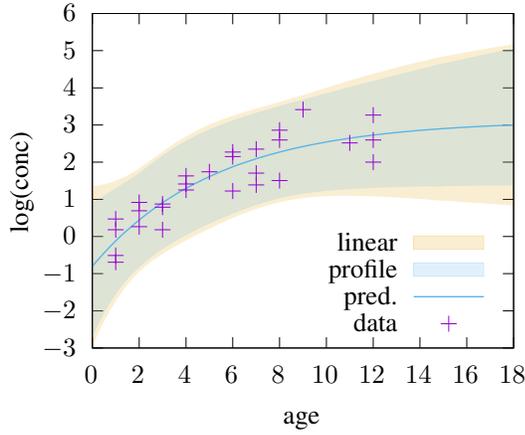}
  \caption{Comparison of linear approximation and likelihood-profile based prediction intervals ($\alpha=0.05$) for the PCB model Eq.~\ref{eqn:hl-logpcb}. Notice that the color for the profile t region blended with the linear prediction interval.}
  \label{fig:prediction-logpcb}
\end{figure}

In Fig.~\ref{fig:prediction-logpcb} we can see that both types of prediction intervals for Eq.~\ref{eqn:hl-logpcb} contain all of the data points. The linear PI is inconsistent with the model for low and high age. In these regions the intervals are not monotonically increasing. Observe that the profile PI correctly matches the monotonicity enforced by the model. The profile-based interval is narrower overall but especially for ages close to zero. Accordingly, the profile-based PI is much better for the PCB model than the linear approximation. 

\subsection{Comparison of SR models for the Kotanchek function}

Next, we use  three different SR algorithms to generate models for the \emph{Kotanchek} function~\cite{Vladislavleva:2009:TEC,smits:2004:GPTP}. The generating function is given by:

\begin{equation}
  f_1(x,y) = \frac{e^{-(x - 1)^2}}{1.2 + (y - 2.5)^2}  \label{eqn:kotanchek}
%   f_2(x,y) &= \frac{(x - 3)^4 + (y-3)^3 - (y-3)}{(y-2)^4 + 10} \label{eqn:ratpol2d} \\
 %    f_3(x,y) &= \frac{1}{1 + x^{-4}} + \frac{1}{1 + y^{-4}} \label{eqn:pagie}
\end{equation}

The training set is generated as described in \cite{Vladislavleva:2009:TEC} by random sampling of 100 observations uniformly from the region $[0.3, 4]^2$. The test set contains $45 \times 45$ points on the regular grid $[-0.2:0.1:4.2]^2$. The test set covers a larger space than the training set which means that extrapolation capability is necessary. 

% \begin{table}
%   \centering
%   \begin{tabular}{lcccc}
%     Function & Partition & Rows & Input space \\
%     \hline
%     $f_1(x,y)$ & train &     100 & $\text{uniform}(0.3, 4)^2$\\
%                & test  & $45*45$ & $(-0.2:0.1:4.2)^2$ \\
%     $f_2(x,y)$ & train &      50 & $\text{uniform}(0.05, 6.05)^2$\\
%                & test  & $34*34$ & $(-0.25, 0.2, 6.35)^2$ \\
%     $f_3(x,y)$ & train & $26*26$ & $(-5:0.4:5)^2$ \\
%                & test  &    1000 & $\text{uniform}(-5,5)^2$\\
%   \end{tabular}
%   \caption{Data generation for the SR benchmark functions.}
%   \label{tab:sampling}
% \end{table}

Table~\ref{tab:models} lists the models found by the three SR algorithms after simplification. The simplified models are produced by a partially automatic process using a computer algebra software with manual intervention to ensure that the expressions have no redundant parameters.

\begin{table*}
   \centering
   \caption{SR models generated by each algorithm for the Kotanchek function.}
   \begin{tabular}{lcc}
     Alg. & $s^2$ & Model \\
     \hline
     HL & $5.8e^{-5}$ & $ \exp(\theta_0 y + \theta_1 x - \theta_2 y^3 -\theta_3 x^2 ) (\theta_4 y +  \theta_5 y^3 + \theta_6)$ \\
% X1 * -1.84086097e-002 + exp((X1)^2f * -1.08709856e+000) * exp(X1 * 2.21285790e+000) * exp(X2 * 2.88777877e+000) * exp((X2)^2f * -5.77217113e-001) * 6.69692943e-003 + 6.53182942e-002
     DSR & $1.55e^{-4}$ & $\theta_0 x + \exp (\theta_1 x^2) \exp(\theta_2 x) \exp(\theta_3 y) \exp(\theta_4 y^2) \theta_5  + \theta_6$ \\
     ITEA & $2.19e^{-4}$ & $\exp \left(\frac{\theta_0 + \theta_1 y + \theta_2 \cos{(x)} + \theta_3 \sin{(x)} + \theta_4 y^2}{1 + \theta_5 x^2}\right)$ \\
   \end{tabular}
   \label{tab:models}
 \end{table*}
% my calculation is 4.61e^{-4} for the ITEA model. Why the difference?

The original expression produced with HL is:
\begin{multline}
  ((\theta_0 y)^2 - \theta_1 y + \theta_2) \\
  \exp(\theta_3 \theta_4 x x + \theta_5 x + \theta_6 y + (\theta_7 y)^3) \theta_8 + \theta_9
\label{eqn:hl-kotanchek-orig}
\end{multline}

This expression is over-parameterized ($\theta_3, \theta_4,\theta_8$) and contains unnecessary nonlinear operations on parameters (e.g.~ $\theta_0^2$).
Algebraic transformation and manual removal of redundant parameters as well parameters which are effectively zero leads to the simplified expression with a new parameter vector:
\begin{multline}
  \exp(\theta_0 y + \theta_1 x + \theta_2 y^3 + \theta_3 x^2) (\theta_4 y +  \theta_5 y^3 + \theta_6) \\
  % PARAMETERS MAY BE REMOVED
  %\theta_0 = 1.9314,\, \theta_1=1.9819,\, \theta_2=0.04764, \theta_3=0.9927,\\
  %\theta_4=-8.554e^{-3},\, \theta_5= 1.818e^{-4},\, \theta_6= 0.02364
  \label{eqn:hl-kotanchek-simplified}
\end{multline}

The maximum likelihood estimate of the parameters with standard errors as well as the parameter correlation matrix are shown in Fig.~\ref{fig:hl-kotanchek-fit}.
\begin{figure}
  \footnotesize
  % TABLE MAY BE REMOVED FOR SPACE
\begin{verbatim}
P  Estimate Std. err. Correlation matrix
0  1.932    3.088e-2  1.00
1  1.982    2.432e-2  0.07  1.00
2  4.764e-2 3.272e-3  0.88  0.10  1.00
3  9.928e-1 9.993e-3  0.05  0.97  0.07  1.00
4 -8.554e-3 4.232e-4  0.95  0.31  0.88  0.27  1.00
5  1.818e-4 8.742e-6 -0.90 -0.31 -0.89 -0.27 -0.99 1.00
6  2.365e-2 1.096e-3 -0.95 -0.33 -0.82 -0.28 -0.99 0.96 1.00
\end{verbatim}
% original table
% Para       Estimate      Std. error          Lower          Upper Correlation matrix
%     0    1.9313e+000    3.0880e-002    1.8700e+000    1.9927e+000 1.00
%     1    1.9819e+000    2.4320e-002    1.9336e+000    2.0302e+000 0.07 1.00
%     2    4.7635e-002    3.2720e-003    4.1137e-002    5.4133e-002 0.88 0.10 1.00
%     3    9.9273e-001    9.9927e-003    9.7289e-001    1.0126e+000 0.05 0.97 0.07 1.00
%     4   -8.5543e-003    4.2319e-004   -9.3946e-003   -7.7139e-003 0.95 0.31 0.88 0.27 1.00
%     5    1.8184e-004    8.7418e-006    1.6449e-004    1.9920e-004 -0.90 -0.31 -0.89 -0.27 -0.99 1.00
%     6    2.3645e-002    1.0956e-003    2.1469e-002    2.5820e-002 -0.95 -0.33 -0.82 -0.28 -0.99 0.96 1.00

  \caption{Maximum likelihood fitting results for Eq.~\ref{eqn:hl-kotanchek-simplified}.}
  \label{fig:hl-kotanchek-fit}
\end{figure}

\begin{figure}
  \centering
  \input{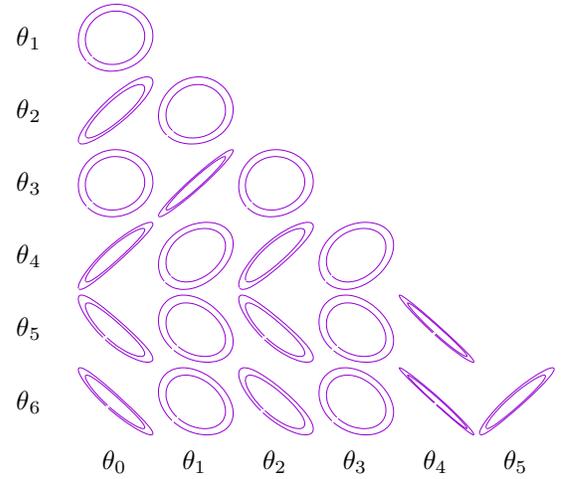}
  \caption{Pairwise profile confidence region plots for Eq.~\ref{eqn:hl-kotanchek-simplified}.}
  \label{fig:hl-pairwise-kotanchek}
\end{figure}

Fig.~\ref{fig:hl-pairwise-kotanchek} shows contour plots for the pairwise confidence regions generated with Alg.~\ref{alg:profilet} and Alg.~\ref{alg:thetathetaplot}. The outer contour represents the $80\%$ confidence region, the inner contour the $50\%$ confidence region. The plots show that several parameters have high correlation. Many contours are close to ellipsoids which indicates that the parameter profiles are close to linear around the maximum likelihood estimate. This implies that the linear approximation is close to the profile-based PI for this model.

\begin{figure}
  \centering
  \input{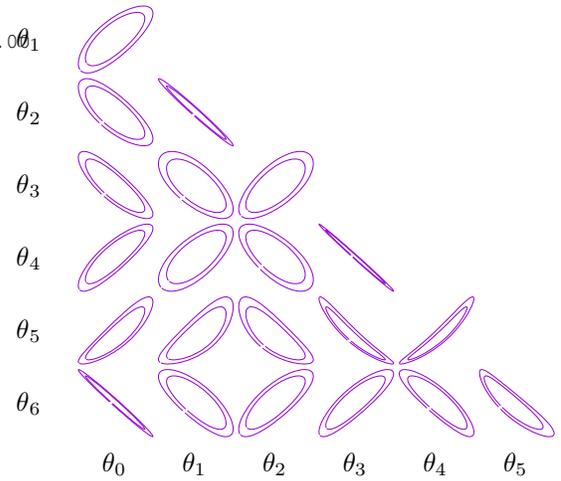}
  \caption{Pairwise profile confidence region plots for the DSR model.}
  \label{fig:dsr-pairwise-kotanchek}
\end{figure}

Fig.~\ref{fig:dsr-pairwise-kotanchek} shows the same plot for the DSR expression. This plot reveals a nonlinear relationship between some of the pairs of parameters. This model not only contains different parameters with nonlinear behavior but also interaction between most of the parameters.
% The thinner ellipses help us to determine more precise intervals at different points of interest.

\begin{figure}[t!]
  \centering
  \input{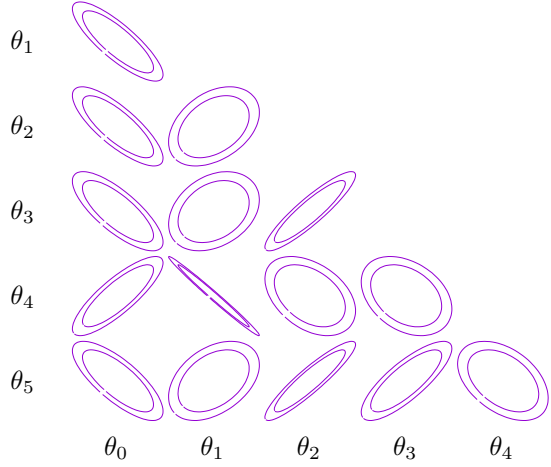}
  \caption{Pairwise profile confidence region plots for TIR model as depicted in Table~\ref{tab:models}.}
  \label{fig:pairwise-kotanchek-tir}
\end{figure}

In Fig.~\ref{fig:pairwise-kotanchek-tir} we can see the same plots for the model created by TIR. This model is an exponential of a rational polynomial. Surprisingly, the uncertainties of this model can be well approximated by the linear approach despite the nonlinearity of the exponential function. We can observe thinner ellipses for terms that uses the same input variable ($\theta_1$ and $\theta_4$, and $\theta_2, \theta_3, \theta_5$) revealing their high correlation. 

%\begin{figure}[t!]
%  \centering
%  \input{figs/TIR_Kotanchek_profile}
%  \caption{Plot of the third parameter of TIR model illustrating a model with a bad behavior at specific points.}
%  \label{fig:pairwise-kotanchek-p3-tir}
%\end{figure}

\begin{figure}
  \centering
  \input{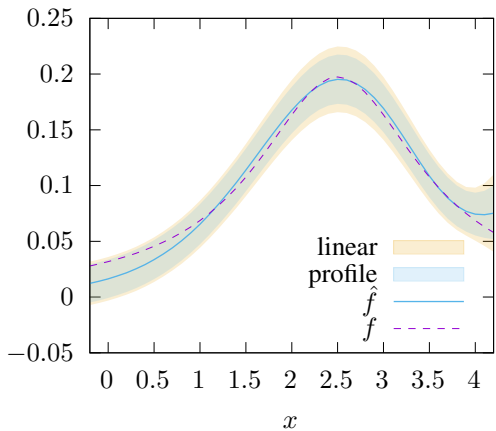}
  \caption{Comparison of linear approximation and likelihood-profile based prediction intervals ($\alpha=0.05$) for Eq.~\ref{eqn:hl-kotanchek-simplified}.}
  \label{fig:prediction-kotanchek}
\end{figure}

Fig.~\ref{fig:prediction-kotanchek} shows a small section from the Kotanchek function for a comparison of the PIs for Eqn.~\ref{eqn:hl-kotanchek-simplified}. The visualization shows that both intervals contain the true function. The profile intervals are narrower which shows the advantage of using profile intervals over the linear approximation even for this model where the parameter are close to linear. % GKR: we used F() for the profile interval and t() for the linear interval.

% DSR

% As we can see in Fig.~\ref{fig:output}.
% 
% \begin{figure*}[t!]
%     \centering
% \begin{lstlisting}[language=bash, frame=single]
% SSR 16.6931 s^2 0.0269
% 
% Corr. Matrix
% [1.   0.74 0.76 0.29]
% [0.74 1.   0.37 0.13]
% [0.76 0.37 1.   0.03]
% [0.29 0.13 0.03 1.  ]
% 
% Linear estimation
% theta	Estimate	Std. Error.	Lower		Upper		
% 0	7.0081e-01	1.7972e-02	6.6551e-01	7.3610e-01	
% 1	4.0342e+01	2.7912e+00	3.4861e+01	4.5823e+01	
% 2	8.7075e+00	1.2744e-01	8.4573e+00	8.9578e+00	
% 3	6.9117e-03	1.1077e-03	4.7364e-03	9.0870e-03	
% 
% 
% Profile t-standard
% theta	Estimate	Std. Error.	Lower		Upper		
% 0	7.0081e-01	1.7972e-02	6.6399e-01	7.3686e-01	
% 1	4.0342e+01	2.7912e+00	5.6343e+02	6.1818e+04
% 2	8.7075e+00	1.2744e-01	8.4633e+00	8.9658e+00
% 3	6.9117e-03	1.1077e-03	4.90975e-03	9.2361e-03
% \end{lstlisting}
% \caption{Example of output of ProfileT tool.}
% \label{fig:output}
% \end{figure*}
% 
% \begin{figure}[t!]
%     \centering
%     \includegraphics[width=0.47\textwidth]{figs/TIR_theta_theta.png}
%     \caption{$\theta-\theta$ (temporary figure)}
%     \label{fig:tir-theta-theta}
% \end{figure}
% 
% \begin{figure}[t!]
%     \centering
%     \includegraphics[width=0.47\textwidth]{figs/TIR_errorbar_0.png}
%     \caption{errorbar for $x$ (temporary figure)}
%     \label{fig:tir-error0}
% \end{figure}
% 
% \begin{figure}[t!]
%     \centering
%     \includegraphics[width=0.47\textwidth]{figs/TIR_errorbar_1.png}
%     \caption{errorbar for $y$ (temporary figure)}
%     \label{fig:tir-error1}
% \end{figure}

\section{Discussion}

From the above examples, we can see how easily CIs and PIs can be calculated for SR models. The main requirement is that the SR model is algebraically simplified and redundant parameters are removed. The linear approximation is easy to calculate and very accurate when model parameters are well-estimated, i.e. for datasets with low noise or the parameter effects are close to linear. For models with strong nonlinearity the likelihood profile can give much better CIs and PIs as demonstrated with the PCB example.

The likelihood profile can also reveal model issues in particular unbounded intervals for parameter estimates. The pairwise plots can visually aid the practitioner to determine the relationship between two parameters.
% and to assert the correct intervals of one parameter given a fixed point of another.

Whenever these plots reveal a problem with the model, we can try to fix it finding a new model, rewriting the expression or refitting the parameters. This can be done manually or computer-aided with computer algebra systems because SR uses a symbolic model representation.
% This same kind of post-analysis and corrections would be impossible in opaque models since we require a symbolic model to reach this goal.

Albeit these advantages, there are still some shortcomings to this approach. One of them is that the optimization of nonlinear models can have multiple solutions, as it was the case for the Kotanchek model obtained by TIR. This is not an easy problem to tackle and it could require an extension using a Bayesian approach or the use of multi-modal optimization algorithms, increasing the computational cost.

The computational cost of the likelihood profile calculation is also another issue, as this requires multiple runs of a nonlinear optimization. This is most critical for the prediction intervals as it usually involves many data points and, each data point may require a number of optimization steps.

% Limitations:
% Only one optimum is considered, multiple local optima possible, extension for multiple optima would need a Bayesian approach.
%
% Combination of prediction intervals from multiple symbolic models. E.g. from an ensemble of models
%
% Non-monotonic dependency of parameters (e.g. sin(theta x))
%
% Performance. 

\section{Conclusion}
\label{sec:conclusion}

In this paper we have proposed the use of likelihood profiles for estimating the uncertainties of the parameters and predictions of  nonlinear symbolic regression models. Unlike the commonly used linear approximation, the likelihood profile returns more accurate intervals especially for models with strong nonlinearity. 

We gave clear pseudo-code for the algorithms originally described by Bates and Watts~\cite{batesnonlinear1988}, whereby we fixed bugs in the original source and additionally describe how the likelihood profiles can be used for prediction intervals, which is only hinted at briefly in the original source.

Together with this proposal, we also provide a freely available Python library implementing the algorithms that can be used together with most SR implementations as long as they generates a valid \emph{sympy} expression.
% This library supports a textual description of the model uncertainties with linear approximation and profile t calculation and the plots of the relationship between the parameters and the confidence interval and the pairwise relationship between the intervals of two parameters.

We demonstrated the use of likelihood profiles with different data sets and algorithms explaining how to read the generated plots and some possible issues when applying this technique on ill-conditioned models produced by many GP systems.
% In some cases it is possible to fix the model to alleviate such issues, as illustrated in one of the examples.

In conclusion, the likelihood profile method is a valuable statistical technique for analysis and validation of SR models that can help the practitioners to extract additional information about the model, inspect the validity, fix ill-conditioning, and understand the limitations of the model.
% This highlights the importance of having a symbolic model in contrast of an opaque model.

For the next steps we will investigate the multi-modality problem, when the nonlinear model can have multiple equally good parameters, such as $\sin{(\theta x)}$. The likelihood profile is only a local approximation around the maximum likelihood estimate of the parameters and does not consider multiple almost equally likely local optima.
Another direction for future research is the possibility of Bayesian model averaging whereby the local uncertainty of SR models is incorporated. 

% Finally, we will constantly work on the current implementation to improve the performance.

% Future work:
% -'global uncertainty' using a mixture distribution (over multiple models) or multiple optima

\section*{Acknowledgment}

This research is funded by Funda\c{c}\~{a}o de Amparo \`{a} Pesquisa do Estado de S\~{a}o Paulo (FAPESP), grant number 2021/12706-1, Josef Ressel Center for Symbolic Regression, Christian Doppler Research Association. 

\bibliographystyle{IEEEtran}
\bibliography{bib} 

%\begin{IEEEbiography}[{\includegraphics[width=1in,height=1.25in,clip,keepaspectratio]{biopictures/fabricio.jpg}}]{F. O. de Franca}
%is a professor of Computer Science at the Federal University of ABC 
%and head of the Heuristics, Analysis and Learning Laboratory (HAL).
%\end{IEEEbiography}
%\begin{IEEEbiography}[{\includegraphics[width=1in,height=1.25in,clip,keepaspectratio]{biopictures/gkronber.png}}]{Gabriel Kronberger}
% is professor for Data Engineering and Business Intelligence at the
% University of Applied Sciences Upper Austria and head of the Josef
% Ressel Center for Symbolic Regression.
%\end{IEEEbiography}

%\vfill


\end{document}